\documentclass[lettersize,journal]{IEEEtran}

\usepackage{amsmath,amsfonts,amssymb}
\usepackage{algorithm}
\usepackage{algorithmic}
\usepackage{array}
\usepackage{booktabs}
\usepackage{multirow}
\usepackage{textcomp}
\usepackage{stfloats}
\usepackage{url}
\usepackage{graphicx}
\usepackage{cite}
\usepackage[table]{xcolor}
\usepackage{pifont}
\usepackage{tikz}

\graphicspath{{figs/}}

\definecolor{ruleblue}{HTML}{1F5F86}
\definecolor{rulebg}  {HTML}{EAF2F8}
\definecolor{pitamber}{HTML}{8A6D1F}
\definecolor{pitbg}   {HTML}{FBF3E6}
\definecolor{tblhdr}  {HTML}{1B3A57}   % deep navy header band
\definecolor{tblband} {HTML}{EAF1F7}   % light blue zebra band
\definecolor{tblrule} {HTML}{1B3A57}
\arrayrulecolor{tblrule}

\newcommand{\hd}[1]{\textcolor{white}{\bfseries #1}}

\DeclareRobustCommand{\hbfull}{\tikz[baseline=-0.52ex]{%
  \fill[tblhdr] (0,0) circle (0.40ex);}}
\DeclareRobustCommand{\hbhalf}{\tikz[baseline=-0.52ex]{%
  \draw[tblhdr,line width=0.28pt] (0,0) circle (0.40ex);
  \fill[tblhdr] (0,0) -- (90:0.40ex) arc (90:270:0.40ex) -- cycle;}}
\DeclareRobustCommand{\hbnone}{\tikz[baseline=-0.52ex]{%
  \draw[tblhdr,line width=0.28pt] (0,0) circle (0.40ex);}}
\newcommand{\yes}{\hbfull}
\newcommand{\no}{\hbnone}
\newcommand{\pmark}{\hbhalf}

\DeclareRobustCommand{\kw}[1]{\textup{\textsc{#1}}}

\newcommand{\calloutbox}[4]{%
  \par\addvspace{5pt}\noindent
  {\setlength{\fboxsep}{5pt}\setlength{\fboxrule}{0.7pt}%
   \fcolorbox{#1}{#2}{%
     \parbox{\dimexpr\columnwidth-2\fboxsep-2\fboxrule\relax}{%
       \footnotesize
       {\bfseries\textcolor{#1}{\ding{228}\,#3}}\par\vspace{1.5pt}#4\par}}}%
  \par\addvspace{5pt}}
\newcommand{\prule}[2]{\calloutbox{ruleblue}{rulebg}{#1}{#2}}
\newcommand{\ppit}[2]{\calloutbox{pitamber}{pitbg}{#1}{#2}}

\newcommand{\Tgt}{\mathcal{T}}
\newcommand{\Vul}{V^{\star}}

\newcommand{\Ind}{\mathbb{I}}
\newcommand{\E}{\mathbb{E}}
\newcommand{\Prob}{\mathbb{P}}

\newcommand{\IFI}{\mathrm{IFI}}
\newcommand{\PPV}{\mathrm{PPV}}

\usepackage[
  colorlinks=true,
  linkcolor=tblhdr,
  citecolor=tblhdr,
  urlcolor=tblhdr,
  filecolor=tblhdr,
  breaklinks=true,
  bookmarks=true,
  bookmarksnumbered=true,
  pdfstartview={FitH},
  pdftitle={Agentic Security: A Systematization of Tools, Failure Modes,
            and Design Laws for LLM-Driven Penetration Testing},
  pdfauthor={Israt Moyeen Noumi; Tarannum Ahmed Nowshin;
             Md. Mehedi Hasan Bhuiyan Nipu; Mohammad Sakib Mahmood;
             M. F. Mridha},
  pdfsubject={Systematization of agentic security: tools, failure modes,
              and design laws},
  pdfkeywords={agentic security, systematization of knowledge, security tooling,
               survey, autonomous penetration testing, AI red-teaming,
               large language model agents, prompt injection, guardrail drift,
               trusted computing base, reproducibility}
]{hyperref}

\begin{document}

% Must precede every other citation so the control entry is written first into
% the .bbl. Suppresses IEEEtran.bst's "------" substitution for a repeated
% author name; prints nothing itself.
\bstctlcite{IEEEtran:BSTcontrol}

\title{Agentic Security: A Systematization of Tools, Failure Modes, and Design Laws for LLM-Driven Penetration Testing}

\author{Israt~Moyeen~Noumi,
        Tarannum~Ahmed~Nowshin,
        Md.~Mehedi~Hasan~Bhuiyan~Nipu,
        Mohammad~Sakib~Mahmood,
        Md. Jakir~Hossain (Senior Member, IEEE), \\
        and~M.~F.~Mridha (Senior Member, IEEE)
\thanks{I.~M.~Noumi is with the Department of Computer Science and Engineering,
Ahsanullah University of Science and Technology, Dhaka, Bangladesh.}
\thanks{T.~A.~Nowshin is with the Department of Computer Science and
Engineering, BRAC University, Dhaka, Bangladesh.}
\thanks{M.~M.~H.~B.~Nipu is with the Department of Electrical and Computer
Engineering, North South University, Dhaka, Bangladesh.}
\thanks{M.~S.~Mahmood is with the Department of Computer Science, Missouri
State University, Springfield, MO 65897 USA.}
\thanks{J~Hossain is with Center for Advanced Analytics, COE for Artificial Intelligence, Faculty of Engineering and Technology, Multimedia University, Melaka 75450, Malaysia}
\thanks{M.~F.~Mridha is with the Department of Computer Science, American
International University-Bangladesh, Dhaka, Bangladesh.}
% \thanks{I.~M.~Noumi, T.~A.~Nowshin, and M.~M.~H.~B.~Nipu contributed equally to
% this work.}
}

% \markboth{IEEE TRANSACTIONS ON DEPENDABLE AND SECURE COMPUTING,~Vol.~XX, No.~X, 2026}%
% {Noumi \MakeLowercase{\textit{et al.}}: Agentic Security: Tools, Failure Modes, and Design Laws}

\maketitle
\thispagestyle{empty}

\begin{abstract}
Agentic security---the use of large-language-model (LLM) agents to plan,
dispatch, and interpret security tooling---has moved in two years from
demonstration to deployed product, and practitioners now entering the field
are rediscovering the same failures in the same order. This paper
systematizes the discipline from hands-on evaluation: we assess ten widely
used static, dynamic, cloud, orchestration, and AI red-teaming tools for use
inside an unattended pipeline, score each on a four-dimensional Integration
Friction Index separating one-time engineering cost from recurring
organisational, legal, and maintenance cost, and catalogue the failure modes
recurring across them. We then show those failures follow from a handful of
quantitative regularities, derived and stated in closed form. Modelling an
agentic security system as stochastic LLM policies wrapped by a deterministic
mediator, we show that long-lived sessions lose resident evidence at a rate
inversely proportional to phase count, so splitting a pipeline into
short-lived sub-agents extends the usable horizon by exactly the compression
ratio between raw evidence and its summary; that a two-stage verdict cascade
multiplies the likelihood ratios of its two scorers and is the only
economical fix for red-team scorer noise, but collapses to one stage once the
scorers' errors correlate; that recording an outcome a scorer could not
evaluate as ``attack failed'' biases every downstream number, and does so on
precisely the most evasive, most severe responses; that routing between a
costly planner-tier model and a cheap worker-tier one is a knapsack problem
whose textbook solution recovers the planner/worker split most teams already
reach by trial and error; and that for a heavy-tailed tool the execution cap
maximising expected value has the closed form $\beta^{\star}=\alpha v/c$.
Last, we argue that scope and budget enforcement cannot be delegated to a
system prompt, because no such enforcement filters what actually executes,
which places the LLM outside the system's trusted computing base by
construction. A platform we designed and built ourselves, \emph{Inspectra},
runs throughout as a worked instantiation, every mechanism labelled shipped,
partial, or planned---including the parts that did not work.
\end{abstract}


% =============================================================================
\section{Introduction}\label{sec:intro}
% =============================================================================

\IEEEPARstart{T}{he} first sign that something was wrong was a report that
disagreed with itself. The scan had started cleanly: the agent enumerated the
repository, dispatched its analysis, exercised the running application, and
began to write. By the time it reached the final section it was summarising
findings that did not match the evidence it had recorded three phases earlier,
and it confidently described files it had never opened. Nothing had crashed.
No exception had been raised. The system had simply run out of the one
resource nobody had thought to meter: its own context window.

That failure, and roughly a dozen others like it, are the subject of this
paper. None of them are exotic. They are the ordinary consequences of building
an autonomous system out of components that were each designed for a different
operator than the one now driving them---security scanners built for a human
at a graphical console, workflow engines built for deterministic business
processes, and language models built to hold a conversation. \emph{Agentic
security} is the emerging discipline of composing these three things into a
system that plans, attacks, verifies, and explains without a human in the
loop, and almost everything that is hard about it lives in the seams between
them.

The discipline is young but no longer speculative: autonomous agents have
exploited one-day vulnerabilities from advisory text
alone~\cite{fang2024oneday}, compromised websites without human
guidance~\cite{fang2024websites}, and driven structured penetration-testing
workflows end to end~\cite{deng2024pentestgpt,happe2023pwnd,xu2024autoattacker}.
Benchmarks now measure offensive
capability~\cite{zhang2025cybench,bhatt2024cyberseceval}, and indirect prompt
injection through content the agent merely reads is an active area with its
own evaluation harnesses~\cite{greshake2023injection,debenedetti2024agentdojo}.
What is missing is not capability evidence but the engineering account: which
properties a system like this must have, why, and what happens quantitatively
when it does not.

This paper closes that gap in two movements: a systematic look at the tools a
newcomer will actually reach for first---ten static-analysis, dynamic-analysis,
cloud-audit, orchestration, and AI red-teaming tools we integrated, scored, and
in several cases replaced---and an argument that the failures those tools
produce are not one-off mistakes but instances of a handful of general
phenomena, each with a closed-form answer once named correctly: a context
window filling up, a detector operating at low prevalence, a routing decision
that is secretly a knapsack, a runtime distribution with a heavy tail. We
attach every lesson to one of these regularities so it outlives the tool
version that motivated it; a survey of tool releases written in 2026 will be
wrong by 2027, but a statement about how fast a fixed-size context window
saturates will not. Every such lesson traces back to a specific, documented
failure from our own integration work, so each formal claim below carries a
concrete empirical anchor rather than resting on the regularity alone.

Throughout, we use \emph{Inspectra}---the agentic penetration-testing
platform we designed, built, and operate, not a third-party system we merely
evaluated---as a worked example rather than as the paper's contribution. Its
numbers in Section~\ref{sec:case} are real configuration values pulled from a
running system, and every
mechanism we mention is explicitly labelled shipped,
partial, or planned. Architecture diagrams in this field routinely describe
components that were designed but never built; saying plainly which of ours
fall into that category is, we think, itself a useful contribution.

The remainder of the paper makes four kinds of claim, gathered here so a
reader can see the shape of the whole argument before working through it.

\begin{itemize}
\item \textbf{A vocabulary for what makes a pipeline agentic}
      (Section~\ref{sec:model}): an agentic security system as a family of
      LLM policies wrapped by a deterministic mediator, with a measurable
      definition of autonomy that separates a genuinely agentic pipeline from
      a scripted one empirically, not by marketing.
\item \textbf{A tool-by-tool account of integration friction}
      (Section~\ref{sec:tools}): ten tools scored on four cost dimensions
      paid by different people on different timescales, which is why the
      tools easiest to run from a command line are so often the hardest to
      fold into an unattended pipeline.
\item \textbf{Four quantitative regularities} (Sections~\ref{sec:context}--\ref{sec:cost}):
      a rate at which long-lived sessions lose resident evidence; a
      likelihood-ratio account of why cascaded verification is the only cheap
      fix for a noisy red-team scorer, and why that fix vanishes once the two
      scorers correlate; a routing rule that turns out to be a textbook
      knapsack; and a closed-form execution cap for any tool with a
      heavy-tailed runtime.
\item \textbf{An argument that the LLM cannot be the last line of defence}
      (Section~\ref{sec:adversary}): scope and budget enforcement are
      properties of what a system executes, not of what a prompt asks a
      model to do, and a prompt-only implementation of either fails exactly
      when the content the agent reads is trying to talk it out of the rule.
\end{itemize}

Every technique here presupposes written, scope-bounded authorisation naming
the target under test; we do not publish attack payloads, jailbreak strings,
or exploit code. Cloud-posture auditing and network reconnaissance are
adjacent disciplines we touch only where the tool boundary matters
(Section~\ref{sec:tools}) and otherwise leave outside the default pipeline,
for reasons given in Section~\ref{sec:adversary}.

% =============================================================================
\section{Four Concerns, One Report}\label{sec:background}
% =============================================================================

Any agentic security system is, in the end, an attempt to automate what a
small penetration-testing team already does, and that team's work divides
naturally into four concerns: static analysis (SAST), which reads source
code without running it; dynamic analysis (DAST), which exercises the
running application; manual-style penetration testing (PT), which chains and
escalates low-severity findings into a demonstrated exploit with a business
story attached; and AI red-teaming (AIRT), which attacks the model embedded
in the product itself. Skipping any one leaves a specific, predictable blind
spot---Table~\ref{tab:concerns} lays out what each concern does and what a
pipeline misses without it.

\begin{table}[!t]
\renewcommand{\arraystretch}{1.18}
\caption{The four concerns an agentic security system must cover.}
\label{tab:concerns}
\centering
\footnotesize
\begin{tabular}{@{}p{0.115\columnwidth}p{0.37\columnwidth}p{0.40\columnwidth}@{}}
\toprule
\rowcolor{tblhdr}
\hd{Concern} & \hd{What it does} & \hd{Cost of skipping it} \\
\midrule
SAST &
Reads source without executing it: injection sinks, weak crypto, hard-coded
secrets, unsafe deserialisation. &
Misses vulnerabilities visible only in code structure, and secrets sitting in
Git history. \\
DAST &
Exercises the \emph{running} application: crawls, authenticates, fuzzes,
observes. &
Never confirms a static finding is reachable; misses auth bypass, IDOR, and
runtime misconfiguration. \\
PT &
Chains and escalates: turns low-severity findings into a demonstrated exploit
with business impact. &
Leaves a wall of severity scores with no narrative for a developer to act on. \\
AIRT &
Attacks the model inside the product: jailbreaks, prompt exfiltration,
indirect injection, tool abuse. &
Leaves an injection surface invisible to a conventional web-application
firewall. \\
\bottomrule
\end{tabular}
\end{table}

The reason agentic security exists as a distinct discipline, rather than as
four separate scanners run back to back, is that a single planning loop can
in principle cover all four concerns in one run and correlate a static sink
with a dynamic confirmation and an AI-layer weakness in the same report. Most
of this paper is an account of why that promise is difficult to deliver in
practice, and Section~\ref{sec:model} opens by making precise what
distinguishes such a loop from a fixed script in the first place.

Three bodies of prior work frame the rest of the paper. On the offensive side,
PentestGPT decomposes penetration testing into reasoning, generation, and
parsing modules specifically to fight context loss~\cite{deng2024pentestgpt};
Happe and Cito report early evidence that LLMs can drive privilege-escalation
workflows~\cite{happe2023pwnd}; and Fang et al.\ show autonomous exploitation
of websites and of one-day vulnerabilities from advisory text
alone~\cite{fang2024websites,fang2024oneday}. Our contribution is
complementary: rather than demonstrate that an agent can attack something, we
work out the discipline that makes the resulting system trustworthy,
budgetable, and reproducible. On the architecture side, the reason-and-act
loop~\cite{yao2023react}, learned tool invocation~\cite{schick2023toolformer},
and multi-agent frameworks~\cite{wu2023autogen} supply the primitives such a
pipeline is built from, while durable-execution
semantics~\cite{burckhardt2021durable} supply a reliability layer agent SDKs
do not; long-context degradation is documented as position-dependent recall
loss~\cite{liu2024lostmiddle,hsieh2024ruler}, and Section~\ref{sec:context}
gives it a rate. On the attack side, indirect prompt
injection~\cite{greshake2023injection}, direct prompt
override~\cite{perez2022ignore}, and multi-turn strategies such as
PAIR~\cite{chao2023pair} and Crescendo~\cite{russinovich2024crescendo}
constitute the catalogue red-team harnesses automate, and defensive practice
is converging on privilege separation rather than better
prompting~\cite{debenedetti2025camel,beurerkellner2025patterns}---the
position Section~\ref{sec:adversary} argues for from first principles.
Governance vocabulary for the AI-layer concern comes from the OWASP LLM
Top~10~\cite{owasp2025llm}.

% =============================================================================
\section{What Makes a Pipeline Agentic}\label{sec:model}
% =============================================================================

We now make the distinction between a scripted pipeline and an agentic one
precise enough to build on. A \emph{target} $\Tgt$ is everything under test:
its source corpus, its reachable network endpoints, any model-backed surfaces
it exposes, and its infrastructure configuration. The target has a ground
truth $\Vul(\Tgt)$---the finite set of genuine, exploitable weaknesses it
actually contains---and a run of any testing system, agentic or not, produces
a set of findings $\hat V$, giving the usual recall $|\hat V\cap\Vul|/|\Vul|$
and precision $|\hat V\cap\Vul|/|\hat V|$.

A system interacts with its target through a \emph{tool algebra}: a set of
invocations $a=(\tau,\theta,u)$ naming a tool $\tau$, its arguments $\theta$,
and its intended target $u$. What separates an agentic system from a
scripted one is not the presence of an LLM but where control lives. We treat
the system as a family of stochastic LLM policies
$\Pi=\{\pi_0,\dots,\pi_k\}$---an orchestrator and its sub-agents---each
mapping an observation history to a new invocation or a terminal artefact,
sitting above three deterministic pieces that are not decorative: a
\emph{mediator} predicate $g$ deciding whether each proposed invocation
actually executes, a canonicalisation step $\Psi$ turning raw traces into a
stable report, and a budget assignment giving every tool a per-call and
per-session ceiling. Section~\ref{sec:cost} shows what the budget buys, and
Section~\ref{sec:adversary} shows why the mediator, not any policy in $\Pi$,
has to be the thing that enforces scope.

Whether a system built this way behaves like a script or like an agent is an
empirical question, and we can give it a number. Let $O$ be the sequence of
observations a run actually produces and $P$ the realised sequence of
invocations---the plan the system in fact executed. Define
\begin{equation}
\alpha \;=\; \frac{I(P;O)}{H(P)} \;\in\;[0,1],
\label{eq:autonomy}
\end{equation}
the fraction of the plan's own uncertainty that is explained by what the
system observed, with $\alpha:=0$ when the plan has no uncertainty to explain.
A pipeline that always runs the same static analyser, then the same spider,
then the same scanner has a constant plan, so $H(P)=0$ and $\alpha=0$
regardless of how sophisticated the LLM inside it is. A system whose plan is
fully determined by what it has already seen attains $\alpha=1$.
Equation~\eqref{eq:autonomy} is estimable directly from an execution log by
comparing plan entropy across repeated runs on the same target versus runs on
varied targets, which turns ``how agentic is this system, really'' from a
marketing question into a measurement. It also tells us where the paper's
recurring cost of autonomy comes from: every increment of $\alpha$ is an
increment of the non-reproducibility we return to in Section~\ref{sec:adversary}.

Coverage across the four concerns of Section~\ref{sec:background} composes in
exactly the way most single-concern product marketing quietly elides. Split
the ground truth by the concern that can discover each weakness, so that
$\Vul=V_{\mathrm{SAST}}\uplus V_{\mathrm{DAST}}\uplus V_{\mathrm{PT}}\uplus
V_{\mathrm{AIRT}}$, and write $\nu_c$ for the share $|V_c|/|\Vul|$ each
concern reaches on its own. A system that only issues invocations inside some
subset $S$ of the four can, by construction, never surface a member of $V_c$
for $c\notin S$ except by accident, so recall is capped at
$\sum_{c\in S}\nu_c$ no matter how good the tools inside $S$ are---which is
also why the PT concern cannot be waved away as ``just a composition of the
other three'': a pipeline that runs SAST and DAST but never chains a static
sink to a dynamic confirmation reports only the uncomposed findings and
misses every vulnerability that only exists once the two are linked.

\begin{figure*}[!t]
\centering
\includegraphics[width=\textwidth]{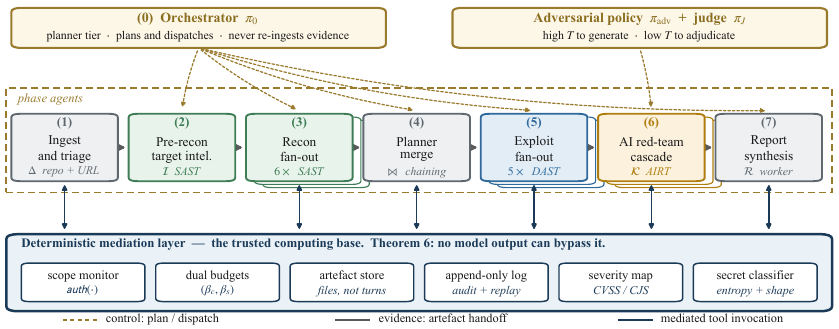}
\caption{Reference architecture, drawn to expose the trust boundary rather
than the data flow. \textbf{(0)} A planner-tier orchestrator plans and
dispatches but never re-ingests raw evidence; a separate adversarial policy
and judge drive AI red-teaming. \textbf{(1)--(4)} Triage, target
intelligence, parallel reconnaissance (stacked cards denote fan-out), and a
planner merge into composite findings. \textbf{(5)--(7)} Exploitation, AI
red-teaming, and report synthesis. Every stage hands its result to the
artefact store rather than a conversation turn (Section~\ref{sec:context});
every invocation is mediated by the deterministic layer beneath, which owns
scope, budgets, severity, and audit (Section~\ref{sec:adversary}).}
\label{fig:arch}
\end{figure*}

Figure~\ref{fig:arch} shows the architecture this section's vocabulary is
meant to describe, drawn deliberately to expose the trust boundary rather than
the data flow. The stochastic layer at the top---the orchestrator and the
adversarial policy that drives AI red-teaming---plans and dispatches, but
never re-ingests the raw evidence it sent sub-agents out to collect; the
short-lived phase agents in the middle do the actual reading and probing and
hand their results to an artefact store rather than back into a conversation;
and the deterministic mediation layer at the bottom owns scope, budgets,
severity, and audit. Section~\ref{sec:context} is about why the middle layer
has to work this way, and Section~\ref{sec:adversary} is about why the bottom
layer cannot be replaced by an instruction in a prompt.

% =============================================================================
\section{The Context Economy}\label{sec:context}
% =============================================================================

\subsection{Saturation}

We return to the report that disagreed with itself, drawn from our own
evaluation log. Once instrumented, the
diagnosis turned out to be unremarkable: the orchestrator had been run as a
single long-lived session, and by the reporting phase the evidence recorded
early in the scan was no longer resident in its context. Nothing was lost in
a way the model itself could detect---from inside a finite window, forgotten
material is indistinguishable from material that never existed. This section
gives that phenomenon a rate, so that a reader can compute in advance how many
phases a given pipeline design can run before it starts to go blind.

Model a monolithic session as a sequence of phases, where phase $i$ appends
$\delta_i>0$ tokens of tool traffic, intermediate reasoning, and retrieved
content to the running conversation. The context window has size $W$ and
carries a static prefix $w_0$ for the system prompt and tool schemas, leaving
an effective capacity $W_{\mathrm{eff}}=W-w_0$. Under the usual
most-recent-first retention that every mainstream provider implements, the
record written during phase $i$ is still resident once phase $n$ executes
exactly when $\sum_{j=i}^{n}\delta_j\le W_{\mathrm{eff}}$. The natural measure
of how much of a run's history is still visible at phase $n$ is its
\emph{retention fidelity},
\begin{equation}
\Phi_n \;=\; \frac{1}{n}\Bigl|\Bigl\{\, i\le n \;:\;
\textstyle\sum_{j=i}^{n}\delta_j \le W_{\mathrm{eff}} \Bigr\}\Bigr| ,
\label{eq:fidelity}
\end{equation}
the fraction of all prior phases whose record has not yet been evicted. If
every phase deposits at least $\delta_{\min}$ tokens, then the set of resident
phases at step $n$ can never hold more than
$\lfloor W_{\mathrm{eff}}/\delta_{\min}\rfloor$ members---each resident phase
contributes at least $\delta_{\min}$ to a total bounded by
$W_{\mathrm{eff}}$---so $\Phi_n$ is squeezed toward zero as $n$ grows,
regardless of how the retention policy is tuned, because any policy that
stores phase records verbatim is subject to the same counting argument. The
first phase whose own record gets evicted, $n^{\star}$, is simply the first
point at which the running total $\sum_{j\le n}\delta_j$ exceeds
$W_{\mathrm{eff}}$; when the per-phase cost is a random variable with finite
mean $\mu$, a standard renewal-theory argument gives
$\E[n^{\star}]\sim W_{\mathrm{eff}}/\mu$ for a large window.

\begin{figure*}[!t]
\centering
\includegraphics[width=\textwidth]{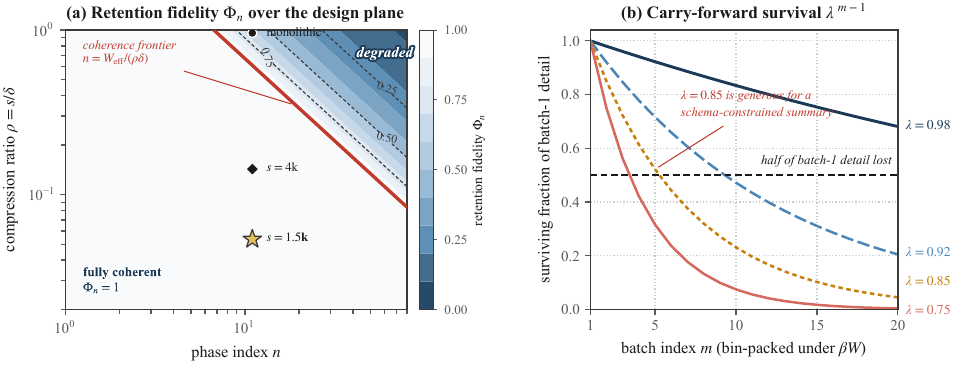}
\caption{The context economy. \textbf{(a)} Retention fidelity $\Phi_n$
of~\eqref{eq:fidelity} over the design plane spanned by the phase count $n$
and the compression ratio $\rho=s/\delta$, for $W=200$k, $w_0=12$k, and
$\delta=28$k. The heavy red curve is the coherence frontier
$n=W_{\mathrm{eff}}/(\rho\delta)$: everything below and to its left retains
all prior phase results, everything above and to its right has begun evicting
them. Moving from a monolithic session ($\rho=1$) to $1.5$k summaries (star)
shifts the frontier right by a factor of $\delta/s$. \textbf{(b)} Geometric
decay of carry-forward detail across bin-packed batches. Both panels are model
predictions under the stated parameters, not measurements.}
\label{fig:context}
\end{figure*}

That asymptotic is the number a practitioner should compute before finalising
a pipeline design, and it is smaller than intuition suggests. With a
$200{,}000$-token window, a $12{,}000$-token static prefix, and phases that
each deposit on the order of $28{,}000$ tokens of tool traffic---a modest
figure once a crawl transcript and a scanner's raw output are
included---$W_{\mathrm{eff}}/\mu$ comes out under seven phases. Figure~\ref{fig:context}(a)
plots retention fidelity across several regimes, and the monolithic curves
fall away almost immediately at the phase counts a real pipeline actually
uses.

\subsection{Hierarchy Buys Exactly the Compression Ratio}

The standard remedy is to give each phase its own short-lived agent, write its
result to an artefact store as a schema-validated summary, and let the
orchestrator ingest only that summary rather than the raw evidence. The gain
this buys is larger than it looks at first glance: if each sub-agent's summary
is capped at $s$ tokens and the orchestrator ingests summaries only, every one
of the first $\lfloor W_{\mathrm{eff}}/s\rfloor$ phases stays fully resident,
and the point at which the pipeline starts losing history is pushed out by
exactly the compression ratio $\rho=s/\delta$ relative to the monolithic case.
This also explains something that puzzled us early on: enlarging the
underlying model's window helped far less than expected, because the raw
traffic $\delta$ grows with the surface being analysed while the summary size
$s$ does not have to.

\prule{Rule~1 --- Pass artefacts, not conversation}{The design lever is the
compression ratio $\rho=s/\delta$, not the window size $W$. Doubling $W$
doubles the horizon; compressing a $28$k-token phase result into a $1.5$k
structured summary multiplies it by nineteen. Short-lived phase agents write
evidence to an artefact store; the orchestrator ingests only schema-validated
summaries and never re-ingests raw evidence---the moment it does, $\rho$
reverts to one.}

\subsection{When a Single Phase Does Not Fit}

Hierarchy fails, though, when a single phase alone already exceeds the
window---most commonly during the first pass over a large repository. The
standard answer is to partition the phase's input into batches under a
conservative ceiling $\beta W$ (with $\beta\approx0.5$, to leave headroom for
the prompt, tool traffic, and reasoning), and to carry a compressed summary
forward from batch $m$ into batch $m+1$. Treating each file's token cost as an
item to be packed, a first-fit-decreasing bin packing comes within a small,
well-understood constant factor of the fewest batches
possible~\cite{dosa2007ffd}, so the number of hops a repository needs is close
to unavoidable rather than an artefact of a bad packer. What matters more than
the batch count, though, is that summarisation across hops is lossy: if each
hop retains only an independent fraction $\lambda\in(0,1]$ of the salient
facts it was given, the expected fraction of the very first batch's detail
still present by batch $m$ decays geometrically as $\lambda^{m-1}$.
Figure~\ref{fig:context}(b) shows the curve, and at $\lambda=0.85$---already a
generous estimate for a schema-constrained summary---half of the first
batch's detail is gone by the fifth hop. Two consequences follow, and both
are less obvious than the decay itself. Packing quality is a \emph{fidelity}
property and not merely a throughput one: fewer hops means less compounding
loss. And any fact that genuinely has to survive to the end of the run should
never be entrusted to the carry-forward channel at all; it belongs in the
artefact store, re-read verbatim at every hop instead of being re-summarised
through it.

Algorithm~\ref{alg:batch} states the resulting procedure. It flags any file
larger than the per-batch payload capacity $\kappa$ as oversized rather than
silently truncating it, and its main loop carries a compressed summary
forward while re-injecting pinned facts verbatim rather than through that
same lossy channel. The validation fallback inside that loop deserves its own
sentence because it is easy to build and easy to skip: a sub-agent
occasionally terminates successfully without writing the output it was asked
to write, and if nothing downstream checks for that, the next stage silently
skips an entire vulnerability
class---a false negative that looks exactly like a clean result. A small
deterministic fallback that notices the missing file and writes an
empty-but-valid record with a diagnostic converts a silent gap into a visible
one. We call this a \emph{code-guarantee save}: non-LLM code repairing a
structural obligation the LLM failed to meet, and one of the few places where
a single line of ordinary code outperforms another layer of prompting.

\begin{algorithm}[!t]
\caption{Saturation-aware batching with carry-forward}
\label{alg:batch}
\begin{algorithmic}[1]
\REQUIRE file set $F$ with token costs $c_f$; window $W$; prefix $w_0$;
         headroom $\beta$; summary cap $s$; pinned facts $\mathcal{P}$
\ENSURE aggregated analysis $A$
\STATE $\kappa \leftarrow \beta W - w_0 - s - |\mathcal{P}|$
\STATE $F' \leftarrow \{f\in F: c_f \le \kappa\}$;
       report $F\setminus F'$ as \kw{oversized}
\STATE sort $F'$ in non-increasing $c_f$ \COMMENT{first-fit-decreasing}
\STATE $\mathcal{B}\leftarrow\emptyset$
\FOR{$f \in F'$}
  \STATE place $f$ in the first $b\in\mathcal{B}$ with residual $\ge c_f$;
         else open a new bin
\ENDFOR
\STATE $\sigma \leftarrow \varnothing$; $A \leftarrow \emptyset$
\FOR{$m = 1$ to $|\mathcal{B}|$}
  \STATE $r_m \leftarrow \pi_{\text{analyse}}(\mathcal{P},\sigma,\mathcal{B}_m)$
         under schema validation
  \IF{$r_m$ fails validation}
     \STATE emit empty-but-valid record with a diagnostic; \textbf{continue}
     \COMMENT{code-guarantee save}
  \ENDIF
  \STATE $A \leftarrow A \cup r_m$;\;
         $\sigma \leftarrow \kw{Compress}(r_m,\sigma)$ truncated to $s$
\ENDFOR
\STATE \textbf{return} $\Psi(A)$ \COMMENT{deterministic canonicalisation}
\end{algorithmic}
\end{algorithm}

% =============================================================================
\section{The Tool Landscape and Integration Friction}\label{sec:tools}
% =============================================================================

\begin{figure*}[!t]
\centering
\includegraphics[width=\textwidth]{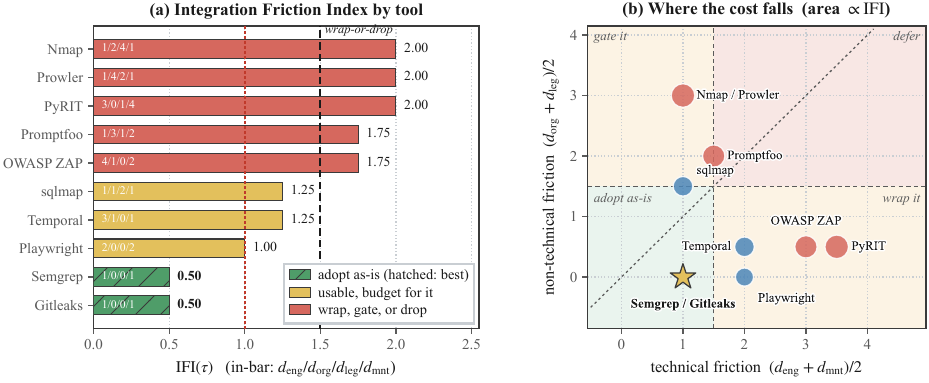}
\caption{Integration friction across the evaluated tool set. \textbf{(a)}
Composite $\IFI(\tau)$ from~\eqref{eq:ifi}, ranked, with the in-bar quadruple
$d_{\mathrm{eng}}/d_{\mathrm{org}}/d_{\mathrm{leg}}/d_{\mathrm{mnt}}$; hatched
bars need no adapter at all. \textbf{(b)} The same data resolved into
technical friction (paid once by engineering) against non-technical friction
(paid per customer by compliance and legal); marker area is proportional to
$\IFI$, and the star is the lowest-friction pair. Most tools sit well off the
equal-cost diagonal, which is why a single scalar rating misleads.}
\label{fig:friction}
\end{figure*}

Tool selection is where most newcomers spend their first month, usually
optimising the wrong variable. The question that matters is not ``how good is
this scanner?'' but ``what does it cost to make this scanner usable by a
process with no human standing next to it?'' The two have different answers
surprisingly often: tools easiest to run from a command line are frequently
the hardest to fold into an unattended pipeline, while a tool costing a week
of integration work can then run untouched for a year. We score every
candidate tool on four independent $0$--$4$ dimensions---the engineering
effort to make it agent-callable, the organisational effort to obtain its
credentials or approvals, the legal authorisation it presupposes, and the
maintenance burden its release cadence imposes---and combine them with
weights on the simplex into a single composite,
\begin{equation}
\IFI(\tau) \;=\; w_{\mathrm{eng}}d_{\mathrm{eng}} + w_{\mathrm{org}}d_{\mathrm{org}}
 + w_{\mathrm{leg}}d_{\mathrm{leg}} + w_{\mathrm{mnt}}d_{\mathrm{mnt}} .
\label{eq:ifi}
\end{equation}
We deliberately do not collapse the four dimensions before reporting them,
because they are paid by different people on different timescales:
engineering cost once by engineering, organisational cost per customer by
sales and compliance, legal cost per engagement by legal, and maintenance
cost forever. Table~\ref{tab:tools} and Figure~\ref{fig:friction} give
uniform-weight scores for the ten tools evaluated; re-weight for your own
organisation and the ranking will change.

\begin{table*}[!t]
\renewcommand{\arraystretch}{1.15}
\caption{Tool landscape with friction decomposition. Scores are
$d_{\mathrm{eng}}/d_{\mathrm{org}}/d_{\mathrm{leg}}/d_{\mathrm{mnt}}$ on
$\{0,\dots,4\}$; $\IFI$ uses uniform weights. \emph{Fit} for an unattended
pipeline: \hbfull~usable as-is, \hbhalf~usable behind a wrapper or an explicit
gate, \hbnone~not usable in the default pipeline. ``Verdict'' records the
disposition in the case study of Section~\ref{sec:case} and the reasoning
behind it.}
\label{tab:tools}
\centering
\scriptsize
\rowcolors{2}{white}{tblband}
\begin{tabular}{@{}p{0.075\textwidth}p{0.098\textwidth}p{0.130\textwidth}p{0.050\textwidth}c c p{0.365\textwidth}@{}}
\toprule
\rowcolor{tblhdr}
\hd{Tool} & \hd{Class} & \hd{Licence / cost} & \hd{Scores} & \hd{$\IFI$} & \hd{Fit} & \hd{Verdict and governing reason} \\
\midrule
Gitleaks~\cite{gitleaks2026}   & Secret scanning   & MIT, free                     & 1/0/0/1 & 0.50 & \yes & Trivial to call; regex false positives need an allow-list from day one. Superseded here by content-level secret scoring. \\
Semgrep~\cite{semgrep2026}    & SAST              & LGPL core, paid cloud         & 1/0/0/1 & 0.50 & \yes & Deterministic JSON output; default rule set is noisy. Complementary to, not replaced by, LLM triage. \\
Playwright~\cite{playwright2026} & Browser automation & Apache 2.0, free              & 2/0/0/2 & 1.00 & \yes & The default answer to authenticated crawling; the work is in making login agent-driven, not in installing it. \\
sqlmap~\cite{sqlmap2026}     & SQLi exploitation & GPLv2, free                   & 1/1/2/1 & 1.25 & \yes & Deterministic exit codes; one of the few tools whose output an agent can trust without an interpretation pass. Needs hard caps. \\
Temporal~\cite{temporal2026}   & Durable workflow  & Apache 2.0 self-host; paid cloud & 3/1/0/1 & 1.25 & \yes & Not a security tool; supplies the durability that agent SDKs do not. Own database and mental model. \\
Promptfoo~\cite{promptfoo2026}  & LLM eval / red-team & OSS core; free tier capped  & 1/3/1/2 & 1.75 & \pmark & Free tier capped at $10^4$ probes/month; heavy catalogue overlap with PyRIT; ownership change is a neutrality consideration. \\
OWASP ZAP  & DAST              & Apache 2.0, free              & 4/1/0/2 & 1.75 & \no & Authentication is a multi-component configuration built around a human at a GUI. See Table~\ref{tab:crawler}. \\
Nmap       & Network recon     & NPSL, free                    & 1/2/4/1 & 2.00 & \no & Technically trivial, legally expensive. Excluded from the default pipeline; opt-in behind a signed scope document. \\
Prowler~\cite{prowler2026}    & Cloud posture     & Apache 2.0, free              & 1/4/2/1 & 2.00 & \pmark & One command to run; weeks to obtain read-only credentials in a governed tenant. On-demand only. \\
PyRIT~\cite{pyrit2026repo}      & AI red-teaming    & MIT, free (inference extra)   & 3/0/1/4 & 2.00 & \pmark & The reference adversarial harness, and marked alpha. Memory-schema and scorer APIs move between minor releases. \\
\bottomrule
\end{tabular}
\end{table*}

One row of Table~\ref{tab:tools} deserves its own explanation, because it is
the single most common surprise for a newcomer: a dynamic scanner runs
successfully, and the entire post-login surface of the application is simply
absent from its site tree. The cause is structural, not a defect in any one
tool. Mainstream open-source dynamic scanners were designed around an
operator who clicks through the login once and hands the resulting session to
the scanner; ZAP presents authentication as a decision tree over context,
method, session management, verification, and user, and a mismatch in any one
of those settings produces traffic that is silently logged out while looking
superficially normal~\cite{zap2026auth}. The direction of travel is
encouraging---Table~\ref{tab:crawler} lays out the options, and ZAP's own
documentation now recommends its Client Spider over the older AJAX Spider for
modern applications~\cite{zap2026ajax}---but the unattended gap remains: the
Client Spider's authentication story is \emph{record once, replay later}, and
a pipeline handed only a repository and a URL has nobody to sit through a
login, nor any guarantee a recording survives the next identity-provider
change.

\begin{table}[!t]
\renewcommand{\arraystretch}{1.18}
\caption{Client Spider versus browser automation for authenticated dynamic
testing; the older AJAX Spider is omitted as ZAP itself no longer recommends
it~\cite{zap2026ajax}. \hbfull~supported, \hbhalf~partial, \hbnone~not
supported.}
\label{tab:crawler}
\centering
\footnotesize
\rowcolors{2}{white}{tblband}
\begin{tabular}{@{}p{0.19\columnwidth}p{0.36\columnwidth}p{0.36\columnwidth}@{}}
\toprule
\rowcolor{tblhdr}
\hd{Property} & \hd{Client Spider} & \hd{Browser automation} \\
\midrule
Fragment routes & \yes & \yes \\
DOM mutations & \yes~streamed live & \yes~probed directly \\
Auth model & Record once, replay & Driven at scan time \\
Unattended? & \pmark~one recording per target & \yes \\
\bottomrule
\end{tabular}
\end{table}

The workable pattern separates discovery from probing: a browser-automation
layer owns the login and emits the authenticated URL set, while the
scanner---or a targeted tool such as sqlmap---consumes it and does what it is
actually good at. The underlying principle applies to nearly every
human-designed tool an agent has to drive, not just to crawlers:
\emph{delegate authentication to whatever layer can hold a browser, and hand
the downstream tool an already-authenticated surface}. It has an honest
limit, though: multi-factor and step-up authentication defeat browser
automation as reliably as they defeat a spider, so a pipeline should say
plainly which authentication modes it supports rather than skip the rest
silently.

Two rows of Table~\ref{tab:tools} score low on engineering and high overall,
and in our experience they are the rows that most often derail a schedule.
Cloud-posture auditing is one command, but ``read-only'' in cloud identity
management is a composition of role grants rarely correct on the first
attempt~\cite{prowler2026}, and the approval cycle routinely outlasts the
scan. Network reconnaissance is technically trivial and legally expensive:
unauthorised port scanning is actionable in many jurisdictions, and scanning
infrastructure a customer does not wholly own can be unauthorised access even
when the customer commissioned the assessment. The right default is the same
in both cases: ship the integration, gate it behind an explicit opt-in, and
never let a first scan wait on a credential that has not arrived.

% =============================================================================
\section{A Verification Calculus for AI Red-Teaming}\label{sec:verify}
% =============================================================================

The second failure arrived looking like a success. The red-team harness
reported a large number of successful jailbreaks against a target endpoint,
and manual inspection showed that most of them were well-formed refusals that
happened to quote the harmful phrase back at us. The scorer had seen the
string, and fired. This section shows that this outcome was not a bug but
arithmetic, and derives what to do about it.

A verdict mechanism---human or automated---is characterised by its
sensitivity $\eta$, the probability it fires on a genuinely successful
attack, and its false-positive rate $\phi$, the probability it fires on a
benign response; their ratio $\Lambda=\eta/\phi$ is the mechanism's positive
likelihood ratio, and it is the single quantity that determines how
trustworthy a positive verdict is. For a prevalence $\pi$ of genuine successes
among all attempts, Bayes' rule gives the resulting precision as
\begin{equation}
\PPV \;=\; \frac{\pi\eta}{\pi\eta+(1-\pi)\phi}
      \;=\; \frac{\Omega\Lambda}{1+\Omega\Lambda},\quad
\Omega=\frac{\pi}{1-\pi} .
\label{eq:ppv}
\end{equation}
The pattern-matching scorers shipped by default with most red-team
harnesses---triggering on tokens such as an affirmative opener or the first
line of an enumerated list---operate at roughly $\eta\approx0.9$,
$\phi\approx0.3$ against modern models, which produce exactly those tokens
constantly in benign replies. At a realistic prevalence of $\pi=0.05$,
Equation~\eqref{eq:ppv} gives $\PPV\approx0.14$: six of every seven reported
jailbreaks are spurious. No amount of prompt engineering repairs that number,
because the underlying problem is that $\Lambda=3$, and $\Lambda$ is a
property of the scorer, not of the prompt fed to it.

\subsection{Cascades Multiply Likelihood Ratios}

\begin{figure*}[!t]
\centering
\includegraphics[width=\textwidth]{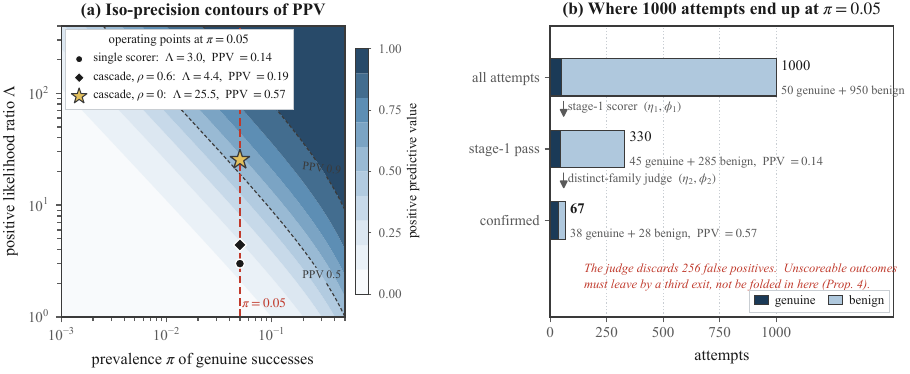}
\caption{The verification calculus. \textbf{(a)} Iso-precision contours
of~\eqref{eq:ppv} over prevalence $\pi$ and likelihood ratio $\Lambda$; the
three markers are the $\pi=0.05$ operating points discussed in the text---a
single heuristic scorer, a correlated cascade ($\rho=0.6$), and an
independent cascade (star). \textbf{(b)} The same result as a population
flow: of $1000$ attempts the stage-1 scorer passes $330$, of which only $45$
are genuine, and the judge removes $256$ false positives. Unscoreable
outcomes leave by a third exit rather than the rejected stream. Model
predictions, not measurements.}
\label{fig:verify}
\end{figure*}

The fix follows directly from the same equation. If a second-stage judge with
its own operating point $(\eta_2,\phi_2)$ is only shown the positives that a
first-stage scorer $(\eta_1,\phi_1)$ already passed, and the two mechanisms err
independently given the true label, then a cascade positive requires both
stages to fire, so the cascade's own sensitivity and false-positive rate are
simply the products $\eta_1\eta_2$ and $\phi_1\phi_2$, and its likelihood
ratio is the product $\Lambda_1\Lambda_2$. Plugging that into the odds form of
Equation~\eqref{eq:ppv} gives
\begin{equation}
\PPV_2 = \frac{\Omega\Lambda_1\Lambda_2}{1+\Omega\Lambda_1\Lambda_2} .
\label{eq:cascade}
\end{equation}
With $(\eta_1,\phi_1)=(0.9,0.3)$, $(\eta_2,\phi_2)=(0.85,0.1)$, and the same
$\pi=0.05$ as above, precision rises from $0.14$ to $0.57$---a fourfold
reduction in the volume of findings a human has to triage, purchased at the
cost of one additional model call per first-stage positive. That trade is
worth making even on pure economics: every spurious positive consumes human
review, which dominates the cost of a scan once inference itself is routed
sensibly, and Figure~\ref{fig:verify}(b) shows that the cascade wins across
the whole prevalence range and wins by the widest margin exactly where
red-teaming actually operates, at low prevalence.

That gain depends on an assumption routinely violated in practice: if the
first-stage scorer and the second-stage judge are drawn from the same model
family and merely prompted differently, they tend to fail on the same inputs,
and the independence the cascade relies on quietly disappears. Letting a
mixing parameter $\rho\in[0,1]$ describe how often the judge, on inputs the
first stage already misclassified, simply behaves like the first stage
rather than judging independently, its effective likelihood ratio
$\Lambda_2(\rho)$ decreases monotonically as $\rho$ grows and collapses to
exactly $1$ at $\rho=1$---the cascade degenerates to its first stage and buys
nothing. The instruction this gives is concrete: the second stage has to
differ from the first along an axis that actually matters---a different
model provider, a different modality of evidence, or a rule-based check on a
property the first-stage model cannot fake. Two prompts to the same
underlying model are not a cascade; they are one detector with extra latency.
A related hazard has no clean statistical fix: a target's response can itself
carry an instruction addressed to the judge, so judges must receive candidate
content as delimited data and never as an instruction, the same
privilege-separation discipline Section~\ref{sec:adversary} applies to the
orchestrator as a whole.

\subsection{The Unscoreable-Collapse Bias}

The most consequential defect we encountered, though, is not scorer noise but
a silent mapping most harnesses apply without documenting it, which we
confirmed in our own evaluation of PyRIT. An attempt a scorer cannot evaluate---an
errored call, a provider-blocked response, an unsupported modality, a hedged
partial compliance---is routinely recorded as ``attack did not succeed,''
and public issue reports against at least one widely used harness document
several distinct paths by which this
happens~\cite{pyrit2026issue2044,pyrit2026pr2083}. Write $u_1$ and $u_0$ for
the probability that an attempt is unscoreable given that it genuinely
succeeded and given that it did not, respectively. If unscoreable outcomes are
folded into negatives, the observed sensitivity and false-positive rate become
$\eta(1-u_1)$ and $\phi(1-u_0)$, so the observed likelihood ratio is biased to
\begin{equation}
\tilde\Lambda = \Lambda\cdot\frac{1-u_1}{1-u_0}\;<\;\Lambda
\quad\text{whenever } u_1>u_0 ,
\label{eq:collapse}
\end{equation}
and that condition is exactly what one expects in practice: responses that
trigger a provider block, arrive truncated, or hedge while partially
complying are disproportionately the ones in which the attack was actually
working, so $u_1$ tends to exceed $u_0$. The bias concentrates on precisely
the most severe findings and makes the target look safer than it
is---the worst possible direction for a security instrument to be wrong in.
The fix is structural rather than statistical: \kw{unscoreable} has to be a
first-class outcome reported as bounded coverage, never silently absorbed
into a negative verdict. Algorithm~\ref{alg:cascade} states the resulting
procedure, distinguishing \kw{confirmed}, \kw{rejected}, and \kw{unscoreable}
throughout rather than collapsing the last into the second.

\begin{algorithm}[!t]
\caption{Two-stage verdict cascade preserving \kw{unscoreable}}
\label{alg:cascade}
\begin{algorithmic}[1]
\REQUIRE attempt transcript $t$; objective $o$; target card $\iota$;
         scorer $S_1$; judge $S_2$ (distinct family); budget $\beta$
\ENSURE verdict in $\{\kw{confirmed},\kw{rejected},\kw{unscoreable}\}$
        with evidence record
\STATE persist $\langle$prompt, response, model ids, parameters,
       content hash$\rangle$
\IF{$t$ errored, was provider-blocked, or is of unsupported modality}
   \STATE \textbf{return} \kw{unscoreable} \COMMENT{never a negative}
\ENDIF
\STATE $v_1 \leftarrow S_1(t)$ \COMMENT{cheap, rule-based, high recall}
\IF{$v_1 = 0$}
   \STATE \textbf{return} \kw{rejected}
\ENDIF
\IF{budget for $S_2$ exhausted}
   \STATE \textbf{return} \kw{unscoreable}
\ENDIF
\STATE $v_2 \leftarrow S_2(\kw{Delimit}(t), o, \iota)$
       \COMMENT{content as data, not instruction}
\IF{$v_2$ abstains or is self-inconsistent across a rubric rewording}
   \STATE \textbf{return} \kw{unscoreable}
\ENDIF
\STATE attach CJS axis scores $(g,b,w,d)$ and rationale
\STATE \textbf{return} \kw{confirmed} if $v_2=1$ else \kw{rejected}
\end{algorithmic}
\end{algorithm}

Precision is not the only weakness of automated red-teaming; relevance is the
other. Most harnesses expose only a narrow channel for describing the
target---typically a short objective string---so generated attacks address a
generic assistant rather than the retrieval-augmented portal or code-review
agent actually in scope, whose interesting surface is its tool exposure, its
retrieved corpus, and its business rules. The workable pattern is a
per-target adversarial persona built from reconnaissance output, combined
with a feedback loop that re-injects what the target has refused and
partially conceded---treating the attacker as an agent in its own right
rather than a stateless prompt generator. That is work a practitioner writes;
no harness we evaluated installs it for free.

% =============================================================================
\section{Cost, Routing, and Budgets}\label{sec:cost}
% =============================================================================

The third failure was financial. An agent handed an open-ended exploitation
tool with no ceiling will use every second it is given, because deep-search
tools are built to keep working as long as they are allowed to. A scan that
normally completed inside an hour once ran for six, and the cause was a single
parameter that a tool had been left to investigate with no reason to stop.

\subsection{Model Routing Is a Knapsack}

Model routing is the more strategic version of the same problem. Assign each
agent role $r$ in the pipeline to one of two model tiers, and let $x_r,y_r$ be
that role's input and output token volumes, $q_{r,m}$ the quality of tier $m$
on role $r$, and $\omega_r$ the role's weight in the pipeline's overall
outcome quality. Upgrading a role from the cheap tier to the costly one buys a
quality gain and costs a price difference,
\begin{equation}
\Delta q_r = \omega_r\!\left(q_{r,\mathrm{hi}}-q_{r,\mathrm{lo}}\right),\quad
\Delta c_r = x_r\Delta p^{\mathrm{in}} + y_r\Delta p^{\mathrm{out}} ,
\label{eq:deltas}
\end{equation}
and their ratio $\theta_r=\Delta q_r/\Delta c_r$ is the role's \emph{routing
efficiency}: the quality bought per dollar spent upgrading it. Choosing which
roles to upgrade under a total budget is, once stated this way, a textbook
$0$--$1$ knapsack problem, and in the multi-tier case a multiple-choice
knapsack~\cite{sinha1979mckp,kellerer2004knapsack}: sorting roles by
non-increasing $\theta_r$ and upgrading down the list until the budget runs
out is optimal for the relaxed problem, and the resulting integral choice is
within a single role's own quality gain of the true optimum.

That observation is worth dwelling on, because most teams reach ``a strong
model for the planner, a cheap model for the workers'' by trial and error and
describe it afterward as a heuristic. The knapsack framing shows it is
actually the optimum of a well-posed allocation problem, and it also predicts
when the heuristic fails: a role with both large token volume \emph{and}
large quality sensitivity---long-context extraction from messy pages is the
usual example---has a middling $\theta_r$ and is exactly where a naive
downgrade silently costs quality. The right discipline is to measure
$\Delta q_r$ per role against a small held-out set and re-check it before any
model change, rather than to assume the split above and stop thinking about
it.

The aggregate version of the lever is easy to state and easy to act on. If a
fraction $f$ of a pipeline's tokens sits on the costly planner tier and the
price ratio between tiers is $r$, then the total spend relative to routing
everything to the planner tier is
\begin{equation}
\frac{C_{\mathrm{routed}}}{C_{\mathrm{uniform}}} = f + \frac{1-f}{r} ,
\label{eq:ratio}
\end{equation}
so at a typical price ratio of $r=5$, halving total spend only requires
holding planner-tier traffic under $37.5\%$ of tokens. Two caveats have bitten
us in practice. Retrying a failed cheap call on the expensive tier quietly
erases the entire saving, so the correct policy on a validation failure is to
log it and skip, not to escalate automatically. And the two tiers differ in
refusal profile as well as in raw capability, so downgrading a role can
change what the pipeline is \emph{willing} to do, not merely how well it does
it.

\subsection{Budgets Are a Reliability Control}

\begin{figure*}[!t]
\centering
\includegraphics[width=\textwidth]{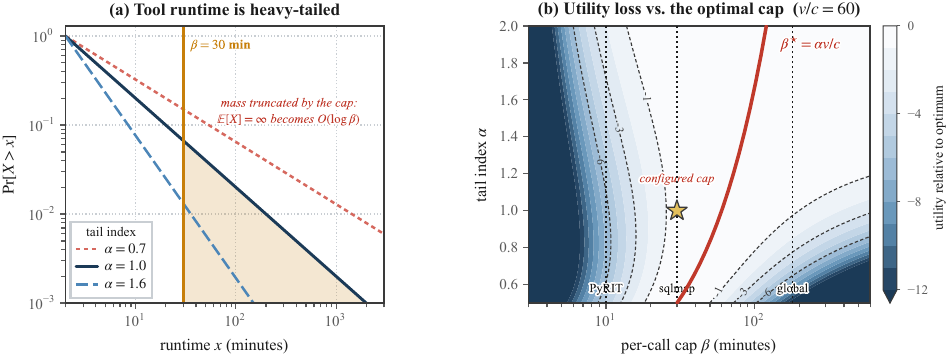}
\caption{Budget economics under heavy-tailed tool runtimes ($x_m=2$~min).
\textbf{(a)} Survival function on log--log axes, where a Pareto tail is a
straight line of slope $-\alpha$; the shaded area right of the cap is the
mass a per-call ceiling removes, turning a divergent expected runtime into
one growing as $\log\beta$. \textbf{(b)} Utility loss relative to the best
attainable cap, over $\beta$ and $\alpha$, at $v/c=60$; the heavy curve is
the optimum $\beta^{\star}=\alpha v/c$ of Equation~\eqref{eq:betastar}, and
the star marks the configured $30$-minute SQL-injection cap. Over-capping is
far cheaper than under-capping, so a generous cap is the safer error when
$\alpha$ is uncertain. Model predictions, not measurements.}
\label{fig:budget}
\end{figure*}

Budgets solve a related but distinct problem. Deep-search security tools have
heavy-tailed runtimes: most invocations finish quickly, but a small fraction
run essentially forever. Modelling the uncapped runtime $X$ of such a tool as
Pareto with scale $x_m$ and shape $\alpha$, so that $\Prob[X>x]=(x_m/x)^{\alpha}$
for $x\ge x_m$, means that for $\alpha\le1$ the expectation itself diverges: an
uncapped tool literally has no finite expected runtime. Suppose a completed
invocation is worth $v$ and running time costs $c$ per unit; the net utility
of capping the tool at $\beta$ is $U(\beta)=v\,\Prob[X\le\beta]-c\,\E[\min(X,\beta)]$,
and a short calculation with the survival function above shows that $U$ is
maximised at
\begin{equation}
\boxed{\;\beta^{\star} = \frac{\alpha v}{c}\;} .
\label{eq:betastar}
\end{equation}
This is, in our experience, the single most immediately usable result in this
paper. It says that the cap should equal the tail's shape parameter times the
ratio of the value of a completed invocation to the cost rate of running
it---both of which a practitioner can actually estimate---and that it depends
on nothing about the tail beyond that one shape parameter. Two structural
facts follow and are visible in Figure~\ref{fig:budget}(a): expected billed
time grows only logarithmically in the cap when $\alpha=1$, so a generous cap
is cheap, and completion probability saturates as $1-x_m/\beta$, so a
generous cap also buys comparatively little beyond a certain point. That
asymmetry is the entire argument for capping in the first place: a modest cap
forfeits a small probability of letting a run finish and removes an unbounded
expectation in exchange. That is a reliability property first and a cost
saving only second, and treating a budget primarily as a cost control leads
teams to set it too high.

A per-call cap alone is not sufficient, because $n$ calls that each
individually respect a cap $\beta_c$ can still aggregate to $n\beta_c$ of
total spend, and a tool-time budget never touches the orchestrator's own
token spend either. All three ceilings below must therefore be enforced
outside the policy being budgeted, by the same mechanism argued for in
Section~\ref{sec:adversary}---and a small reserve should be withheld
deliberately, so an exhausted scan still reports its own coverage limits
rather than simply stopping mid-log. A partial scan reported as if it were
complete is a false negative wearing the costume of a clean result.

\prule{Rule~2 --- Budget twice, then bound the turns}{Set the per-call cap
from $\beta^{\star}=\alpha v/c$, bound the aggregate with a per-session
ceiling, and cap orchestrator turns separately: the three quantities measure
different things and no one of them implies the others. Enforce all three
outside the policy, reserve five to ten per cent of the session budget for
reporting, and label every class the budget truncated as \kw{bounded-by-time}.}

% =============================================================================
\section{The Adversarial Environment}\label{sec:adversary}
% =============================================================================

An agentic security system operates in an environment that is hostile in
three distinct directions at once, and it pays to keep them separate because
each has a different remedy. The target is adversarial by assumption. The
content the system reads is attacker-controlled, so the system is itself a
prompt-injection target the moment it opens a file or renders a page. And the
model provider behind the whole pipeline is a third party whose own safety
posture can change without notice under commercial and regulatory pressure.

Take scope enforcement first, because it is the cleanest case and the one
whose violation is easiest to overlook. Let $\mathcal{D}$ be the authorised
scope for a scan, and let the mediator evaluate a total predicate
$g(a)=\Ind[\mathrm{target}(a)\in\mathcal{D}]$ on every proposed invocation
$a$, executing it only when $g(a)=1$. Because $g$ is evaluated outside every
policy and does not depend on which one produced $a$, the probability that an
\emph{executed} invocation ever violates scope is identically zero for every
policy family---including one fully compromised by an adversary, since a
compromised policy can only propose invocations, and every proposal still
passes through the same predicate. Asserting scope only inside a system
prompt gives no such guarantee: a prompt constraint biases what a policy
tends to output, it does not filter what actually executes, so the violation
probability is at least the probability the policy emits an out-of-scope
invocation, which is in turn lower-bounded by the success rate of prompt
injection against it. This is the classical reference-monitor argument from
operating-system security~\cite{saltzer1975protection,anderson1972rm}, applied
to an LLM agent instead of a process: the system's trusted computing base is
the mediation layer---scope predicate, budget accountant, artefact store,
audit log, canonical severity mapping---and it excludes every LLM policy by
construction. No security property this system needs can be enforced by a
prompt.

That statement is the single most important design claim in this paper, and
it is routinely violated in practice. An instruction such as ``only test the
authorised host'' or ``do not spend more than one hour,'' placed in a system
prompt, is a request rather than a control, and it fails under three
independent conditions: ordinary sampling variance, the context saturation of
Section~\ref{sec:context}, and instructions injected through the content the
scanner is paid to read. That last failure is not hypothetical for a scanner
whose job is to read attacker-controlled material: a repository can contain a
comment addressed to the agent reading it; a crawled page can carry
instructions in its DOM; a document ingested to test for indirect injection
is, by construction, an instruction-bearing artefact. Current defensive
practice is converging on privilege
separation~\cite{debenedetti2025camel,beurerkellner2025patterns,greshake2023injection}:
the component reading untrusted content is granted no network egress and no
tool authority, and its output crosses into the planning path only as
delimited data, exactly as candidate responses cross into a red-team judge in
Section~\ref{sec:verify}. Figure~\ref{fig:arch}'s deterministic mediation
layer is where every such check actually lives---scope monitor, dual-budget
accountant, artefact store, audit log, severity map, secret classifier---each
invoked on every tool call regardless of which policy requested it.
Algorithm~\ref{alg:mediate} states the invocation path that layer follows on
every single tool call: the scope check comes first and unconditionally,
before budget accounting, argument sanitisation, or execution, so that no
downstream step ever runs against a target outside $\mathcal{D}$.

\begin{algorithm}[!t]
\caption{Mediated tool invocation}
\label{alg:mediate}
\begin{algorithmic}[1]
\REQUIRE requested invocation $a=(\tau,\theta,u)$; scope $\mathcal{D}$;
         budgets $(\beta_c,\beta_s)$; consumed $\gamma_\tau$; audit log $L$
\ENSURE result or a typed refusal; invariants preserved
\STATE append $\langle$request, $a$, timestamp$\rangle$ to $L$
\IF{$u \notin \mathcal{D}$}
   \STATE \textbf{return} \kw{out-of-scope} \COMMENT{Section~\ref{sec:adversary}}
\ENDIF
\IF{$\gamma_\tau \ge \beta_s$}
   \STATE \textbf{return} \kw{session-budget-exhausted}
\ENDIF
\STATE $\theta \leftarrow \kw{Sanitise}(\theta)$;
       reject arguments encoding a second target
\STATE $\beta \leftarrow \min(\beta_c,\ \beta_s-\gamma_\tau)$
\STATE $(\text{res},\Delta t) \leftarrow \kw{RunWithDeadline}(\tau,\theta,u,\beta)$
\STATE $\gamma_\tau \leftarrow \gamma_\tau + \Delta t$
\STATE append $\langle$result digest, $\Delta t$, $\gamma_\tau$$\rangle$ to $L$
\IF{$\Delta t \ge \beta$}
   \STATE mark the corresponding coverage class \kw{bounded-by-time}
\ENDIF
\STATE \textbf{return} res as \emph{delimited data}, never as instruction
\end{algorithmic}
\end{algorithm}

\ppit{Pitfall --- A prompt is a request, not a control}{``Only test the
authorised host'' or ``do not exceed one hour'' in a system prompt fails
under three independent conditions: sampling variance, context saturation,
and instructions injected by the very content the scanner is paid to read. If
a property matters, it must be a predicate in the mediation layer---not a
paragraph in a prompt.}

\subsection{Guardrail Drift, and Why It Is Not the Vendor's Fault}

The second front is the model provider itself, and the failure here was
less abstract. Mid-project, the component composing adversarial prompts for
authorised testing began refusing work it had performed the week before; no
code had changed, the provider's safety classifier had. The publicly
documented mid-2026 sequence is fully attributable: a frontier model shipped
in a safeguarded and a restricted-access variant; an export-control
directive followed three days later and, since nationality cannot be
verified in real time, access was suspended globally; the vendor spent the
following weeks building an improved classifier after an externally reported
jailbreak; and controls lifted after eighteen
days~\cite{anthropic2026redeploy,anthropic2026mythos}. The vendor's own
account states the technique is now blocked in over $99\%$ of cases but that
the change ``comes at the cost of flagging benign requests more often during
routine coding and debugging tasks''~\cite{anthropic2026redeploy}, and its
published use taxonomy places penetration testing, red teaming, and bug
bounties in a high-risk dual-use category blocked outright pending better
means of identifying authorised users~\cite{anthropic2026safeguards}. The
flagship use case of a provider-driven pentest agent is, on the most
safeguarded tier, out of scope by policy.

That second half of the quotation is structural, not an implementation
detail: moving a decision threshold to catch more genuine misuse necessarily
catches more legitimate traffic too, wherever the legitimate-request
distribution has density near that threshold. Only retraining the scorer
escapes the trade-off, and even then the vendor reported both a higher catch
rate \emph{and} more benign flagging, suggesting the operating point simply
moved along the curve. Either way, the practitioner's conclusion is the
same: every legitimate defensive user pays a false-refusal tax whenever a
provider tightens its threshold, and that tax must be engineered around, not
argued with.

The responses that survive contact with this are unglamorous, and mostly
amount to reducing exposure rather than negotiating with it. Pin dated model
snapshots rather than floating aliases. Move payload \emph{authorship} out of
the safeguarded model entirely, templating offensive strings in deterministic
code and confining the model to scope framing and summarisation. Declare
authorised scope explicitly in every system prompt---it costs nothing, and it
matches the vendor's own description of benign use. Run a behavioural canary
suite at the start of every campaign, so a classifier shift is caught before
it is mistaken for a target that has become secure. Record, per scan, the
acceptable-use policy version in force and the authorisation basis for the
scan~\cite{anthropic2025aup}; that is compliance evidence, not telemetry. And
keep the provider abstraction thin enough that swapping providers is
configuration, not a rewrite: if unavailability events across independent
providers are themselves independent with per-provider probability $q$, path
availability across $n$ providers is $A=1-\prod_{i\le n}q_i$. An eighteen-day
annual outage gives $q\approx0.049$, so one provider buys about $95.1\%$
availability while a second raises that to roughly $99.76\%$---independence
is optimistic, since one regulatory action can bind several vendors at once,
but the first additional provider is still worth an order of magnitude in
downtime.

A boolean jailbreak verdict is nearly useless on its own: it cannot
distinguish a technique that merely reproduces a public tutorial from one
that confers genuine new capability. The Cyber Jailbreak Severity draft
addresses this with four axes on $[0,4]$---capability gain, breadth, ease of
weaponisation, and discoverability---combined into a band from CJS-0
(informational, no gain over public tooling) through CJS-4 (critical: broad,
easily weaponised, readily discoverable), with capability gain gating the
rest, so zero uplift scores as purely informational regardless of ease of
weaponisation~\cite{anthropic2026safeguards}. We recommend axis scoring of
this kind regardless of whether the draft is ratified, for the same reason
CVSS is worth using for conventional vulnerabilities: a shared vocabulary
makes findings comparable in language regulators and vendors already use.
The instrument should also be turned on itself---a tool that can itself be
prompted into authoring working exploits has a severity score of its own,
and the field would benefit from publishing them routinely.

\subsection{Reproducibility}

Two customers independently reported that re-running the same scan against
unchanged code produced different finding counts, with neither the target
nor the model version having changed. The variance came from ordinary
sampling, and it is worth being precise about why tuning temperature alone
cannot eliminate it. Every sampled decision an LLM policy makes is, in
effect, a small biased coin flip, and two independent runs can only follow
an identical trajectory if every one of those flips lands the same way
twice; if $L$ is the number of such decisions and $\bar\kappa<1$ the largest
probability that any single decision recurs identically, the probability
that two full runs agree on every decision is at most $\bar\kappa^{\,L}$---so
the count of sampled decisions matters exponentially more than how sharply
any one is distributed. Lowering temperature reduces $\bar\kappa$ a little;
removing a decision from the sampler entirely shrinks the exponent, a
categorically larger effect. That is why the highest-yield interventions are
structural: classify files by rule so the input to every later phase is
stable, bound turn counts so exploration cannot diverge, and require
schema-validated output on every internal call so harmless prose variation
cannot propagate into a decision that should not depend on it. Lowering
temperature on planning and tool-selection roles still helps---the common
default of $1.0$ suits a role that should choose the same tool given the
same evidence poorly---but it is a second-order lever next to the ones above.

What is genuinely attainable, even though byte-for-byte trajectory agreement
is not, is \emph{semantic} stability: deduplicate findings by content hash,
sort by a stable key, and map severity through a lookup table rather than
another model call. Because that canonicalisation is many-to-one, two runs
can agree closely on the resulting report even when their trajectories agree
on almost nothing, and the operationally meaningful metric becomes set
agreement over the discovered surface,
$|\hat V_1\cap\hat V_2|/|\hat V_1\cup\hat V_2|$, together with byte-level
replay of one recorded transcript for audit. State plainly which kind of
reproducibility a system actually offers; conflating the two is a fast way
to lose a customer's trust the first time the numbers move.

% =============================================================================
\section{Case Study: Inspectra}\label{sec:case}
% =============================================================================

Inspectra is the agentic security platform we designed, built, and
operate---not a system we merely evaluated. It combines static analysis,
authenticated dynamic analysis, and AI red-teaming behind an LLM
orchestrator, distributed for customer-hosted deployment rather than as a
hosted service, and every finding tagged by number throughout this paper
comes from building it. We use Inspectra here as a worked instantiation and
label every mechanism below as shipped, partial, or planned, because the gap
between design intent and running code is exactly what a reader calibrating
their own schedule most needs to see.

\subsection{Instantiation}

Every scan in Inspectra is a durable workflow, and every phase within it is an
activity, which supplies the retries, timers, cancellation, and restart
survival that an agent SDK alone does not
provide~\cite{temporal2026,burckhardt2021durable}. Deterministic triage
classifies files by rule before any model call touches them, emitting a
stable file map---the same reproducibility intervention argued for in
Section~\ref{sec:adversary}. Target intelligence and six per-class
reconnaissance agents run on the planner tier; five per-class exploitation
agents, the report agent, and the red-team dispatcher run on the worker tier,
which is the routing rule of Section~\ref{sec:cost} realised directly as
configuration rather than left as an aspiration. Authenticated discovery is
browser-driven, taking a credential or token at scan time, and the resulting
URL set feeds sqlmap through a mediated tool call. Reports are emitted as
Markdown so a downstream repair agent can consume them without a lossy
conversion step. Table~\ref{tab:config} gives the resulting configuration.

\begin{table}[!t]
\renewcommand{\arraystretch}{1.12}
\caption{Inspectra configuration. Budgets are enforced in the mediation layer,
outside policy control.}
\label{tab:config}
\centering
\footnotesize
\begin{tabular}{@{}lccl@{}}
\toprule
\rowcolor{tblhdr}
\hd{Role / tool} & \hd{$\beta_c$} & \hd{$\tau_{\max}$} & \hd{Tier} \\
\midrule
Orchestrator                 & 180\,min (global) & 50 & planner \\
Pre-recon (target intel.)    & 30\,min & 60 & planner \\
Recon, per class ($\times6$) & 30\,min & 60 & planner \\
Exploit, per class ($\times5$) & 30\,min & 15 & worker \\
SQL-injection tool           & 30\,min & --- & --- \\
AI red-team, per attack call & 10\,min & --- & --- \\
AI red-team agent envelope   & 40\,min & 60 & worker \\
Report synthesis             & 30\,min & 40 & worker \\
Repair / re-verification     & --- & 100 & worker \\
Per-session ceiling, all tools & 60\,min & --- & --- \\
Batching ceiling             & \multicolumn{3}{l}{$10^5$ tokens ($\beta\approx0.5$ of a $2\times10^5$ window)} \\
\bottomrule
\end{tabular}
\end{table}

\subsection{What Is Shipped, What Is Partial, What Is Planned}

What is actually shipped includes hierarchical orchestration with
artefact-based handoff; the mediation layer with dual budgets and per-run
event logging; deterministic triage and file mapping; browser-driven
authenticated discovery; a nine-strategy red-team dispatcher; deterministic
severity mapping with a model fallback; content-level secret scoring that
demotes obvious placeholders and promotes high-entropy or provider-shaped
credentials; synthetic environment generation so a repair agent can boot a
patched project without production secrets; encrypted prompt bundles with a
signed licence gate; and the minimal code-guarantee save of
Section~\ref{sec:context}. What is only partial is more instructive: the
judge stage of our verdict cascade is currently a heuristic scorer rather
than a distinct-family LLM judge, so the precision gain of
Section~\ref{sec:verify} is not yet realised; the batching module is
implemented and unit-tested but not wired into the reconnaissance runner,
because the single-batch path has sufficed for every repository tested so
far; and temperature is not pinned per role, with the resulting variance
accepted in exchange for exploratory breadth and determinism recovered
structurally instead. Some things are only planned. A two-branch
environment-file rule---no finding when a file is properly ignored, critical
when it is not---is not yet a discrete gate: ignore rules are
already parsed and every file is stamped with the resulting flag, but nothing
downstream branches on it yet, so real leaks still surface only through
content-level scoring. A fuller code-guarantee
save that salvages content directly from a transcript is not implemented, and
cloud-posture auditing is available on request rather than wired in. We
record all three gaps deliberately: secondary judges, code-guarantee saves,
and complete cloud coverage are precisely the mechanisms that tend to reach
an architecture diagram long before they reach running code.

Three further findings from that same integration log recur often enough
across projects like this one to be worth recording. Pinning the red-team
dependency to Python 3.12 was, for most of the project, a hard requirement:
releases through 0.13.x narrowed PyRIT's supported interpreter versions,
forcing the backend onto one Python minor, and only 0.14 widened that back
out---an alpha-status dependency's risk is not only in its API
but in the runtime it insists on. A healing agent that verifies a fix by
running the patched project needs a real interpreter, but triage has
correctly already excluded the customer's own \texttt{venv/} or
\texttt{node\_modules/}; the fix is a fresh, disposable environment built
from the project's own manifest for every healing attempt, never a reuse of
anything the customer shipped. And a customer who zips a
project with its virtual environment included makes triage attempt to
analyse tens of thousands of vendored files before reaching their own code,
so the practical fix is to exclude well-known dependency directories by
name and document the exclusion.

\subsection{Deployment Economics}

The deployment story mattered as much as any algorithm above. The stack is
multi-container by necessity---an interface, a front end, a shared database, a
workflow server, and several tool runners---so single-container platform
offerings cannot host it at all. Managed Kubernetes is the lowest-effort
answer, with a floor we estimate around \$280 per month; that figure is our
own directional estimate, not a quotation. Platform support here is itself a
moving target: the compose-based path on one major cloud provider already
carries a published 2027 retirement notice with migration to sidecars or
managed Kubernetes~\cite{azure2026container}. Our response has been to keep
the tool-runner contract on loopback HTTP, so the same images run unmodified
under sidecars, Kubernetes pods, or plain compose on a customer's own
hardware. Software cost for the tooling itself is zero; compute, storage, and
inference are where the money goes, which is exactly why the routing rule of
Section~\ref{sec:cost} dominates every other lever available.

Customer-hosted deployment resolves the single most common enterprise
objection we encountered---source code must not leave the
perimeter---but it creates a mirror-image problem for us as the
vendor, since our own prompts and orchestration logic now sit on customer
hardware. Encrypted bundles and a signed licence gate raise the
cost of extraction, but Python is not meaningfully obfuscatable and container
layers unpack trivially, so the durable protection here is contractual
rather than technical; a product differentiated by design quality rather
than secrecy is better placed under this constraint than one whose value
sits in a single prompt file. Inference remains the last mile: an air-gapped
customer must accept either an egress proxy or the cost of self-hosting
open-weights models.

For a reader assembling their own pipeline, the shortest version of this
paper's advice fits in five lines: pass artefacts between phases, never raw
conversation; budget every tool twice, per call and per session, and cap
orchestrator turns separately from both; treat the mediation layer, not any
model, as the only thing that may enforce scope or spend; cascade your
red-team verdicts through a genuinely distinct second scorer, and give
unscoreable outcomes their own exit rather than folding them into a rejection;
and route model tiers by the quality each role actually buys per dollar,
re-measured, rather than by convention.

% =============================================================================
\section{Discussion}\label{sec:discussion}
% =============================================================================

Every quantitative claim here is analytic, and its force depends on
assumptions worth stating plainly. Context saturation
(Section~\ref{sec:context}) assumes most-recent-first eviction; real
providers implement attention and caching in ways that make ``resident'' a
softer notion than the model allows, and positional-recall studies suggest
degradation begins well before formal
eviction~\cite{liu2024lostmiddle,hsieh2024ruler}, which if anything makes
our bound optimistic. The budget argument (Section~\ref{sec:cost}) assumes a
Pareto runtime; heavy tails in computing workloads are well
documented~\cite{crovella1997heavytail}, but the
shape parameter must be estimated per tool. The cascade argument
(Section~\ref{sec:verify}) assumes conditionally independent errors, and we
quantify the degradation once that fails rather than assuming it holds.
Operating points throughout Sections~\ref{sec:verify}
and~\ref{sec:cost} are illustrative, not measured: we report no precision or
recall figures for Inspectra itself, because no labelled ground truth exists
for its real targets. The Integration Friction Index is an ordinal
instrument scored by the authors, reproducible in method rather than value.
And the vendor-policy timeline of Section~\ref{sec:adversary} is a single
episode; what we generalise from it is only that provider policy is a
first-class engineering dependency, not the episode itself.

Several problems resisted every approach we tried, and a paper reporting
only its successes miscalibrates its readers. Recognising that a static sink
and a dynamically confirmed exploit are one vulnerability, not two, needs an
identifier minted at first discovery and carried faithfully through later
phases; agents invent new identifiers or drift on convention, and content
hashing mitigates the problem without closing it. A verifier cannot
adjudicate a success requiring more capability than it possesses, so as
targets improve, a cascade's judge must improve with them or its economics
degrade. No labelled corpus of attack-succeeded-versus-refused generalises
across targets, so precision claims here---ours included---are not
comparable between systems. A target can be crafted to maximise agent
deliberation through cheap-looking calls, consuming a session budget without
producing a finding, and rate-shaping mitigates this without solving it.
Frontier capability is increasingly confined to vetted participants in
restricted-access programmes~\cite{anthropic2026glasswing}, so any claim
that a public-tool pipeline finds what frontier models cannot should be
checked against the tier compared. And without on-premises inference an
air-gapped deployment cannot reason at all; self-hosting open-weights models
is the only complete answer, and it carries costs most customers have not
budgeted.

Three directions seem most worth pursuing. A standardised \emph{target
descriptor} schema for AI red-teaming would let per-target attack context
transfer between tools and finally make cross-system ablation possible.
Calibrated verdicts rather than binary labels would help close the
unscoreable-collapse problem of Section~\ref{sec:verify}: scorers emitting
posterior probabilities with published operating points, evaluated for
invariance under rubric rewording. And formal verification of the mediation
layer looks genuinely tractable, because Section~\ref{sec:adversary} reduces
the trusted computing base to a small body of deterministic code, small
enough to be within reach of machine-checked proof---a qualitatively
stronger guarantee than anything the
field currently offers, and one that depends on nothing about the
underlying models.

% =============================================================================
\section{Conclusion}\label{sec:conclusion}
% =============================================================================

Agentic security is not, on this evidence, blocked primarily by model
capability. It is blocked by three constraints that are invisible from a
demonstration and unavoidable in production: the alpha-quality of the most
useful adversarial dependencies, the volatility of provider safety policy, and
the cost gradient between model tiers and deployment topologies. Each of the
three now has a precise form rather than a description. Context saturation
sets the usable phase horizon at $W_{\mathrm{eff}}/\delta$, and splitting a
pipeline into short-lived phases multiplies that horizon by the compression
ratio $\delta/s$. Verdict precision at low prevalence is governed entirely by
the likelihood ratio; a genuine cascade multiplies it, and correlation between
scorers destroys the gain. Model routing is a knapsack whose greedy solution
recovers the planner/worker split most teams already reach empirically. The
optimal execution cap for a heavy-tailed tool is $\alpha v/c$. And scope and
budget enforcement are reference-monitor properties that no prompt can
supply, which places every LLM policy outside the system's own trusted
computing base, by construction rather than by convention.

That last claim is the one we would most like a reader to carry away from
this paper. An agentic security system reads attacker-controlled material as
its core function, and then dispatches real tools against real infrastructure
on the strength of what it read. The only defensible architecture for a
system like that is one in which the model proposes and deterministic code
disposes---where scope, budgets, severity, and audit live in a small, boring,
verifiable layer that no output from any model can talk its way past.
Everything else in this paper is engineering discipline built on top of that
one structural decision.

We began with a report that disagreed with itself. A paper like this one does
not prevent the first such report a new team will encounter; what it can do
is shorten the distance between hitting that failure and recognising what
caused it, from a month of debugging down to an afternoon of reading.

% -----------------------------------------------------------------------
% Bibliography. Build with: pdflatex -> bibtex -> pdflatex -> pdflatex
% IEEEtran.bst ships with TeX Live and MiKTeX; no extra download needed.
% -----------------------------------------------------------------------
\bibliographystyle{IEEEtran}
\bibliography{references}

\end{document}